# SceneCut: Joint Geometric and Object Segmentation for Indoor Scenes

Trung T. Pham[1,2], Thanh-Toan Do[1,2], Niko Sünderhauf[1,3], Ian Reid[1,2]

*Abstract*— This paper presents SceneCut, a novel approach to *jointly* discover previously unseen objects and non-object surfaces using a single RGB-D image. SceneCut's joint reasoning over scene semantics and geometry allows a robot to detect and segment object instances in complex scenes where modern deep learning-based methods either fail to separate object *instances*, or fail to detect objects that were not seen during training. SceneCut automatically decomposes a scene into meaningful regions which either represent objects or scene surfaces. The decomposition is qualified by an unified energy function over objectness and geometric fitting. We show how this energy function can be optimized efficiently by utilizing hierarchical segmentation trees. Moreover, we leverage a pre-trained convolutional *oriented boundary* network to predict accurate boundaries from images, which are used to construct high-quality region hierarchies. We evaluate SceneCut on several different indoor environments, and the results show that SceneCut significantly outperforms all the existing methods.

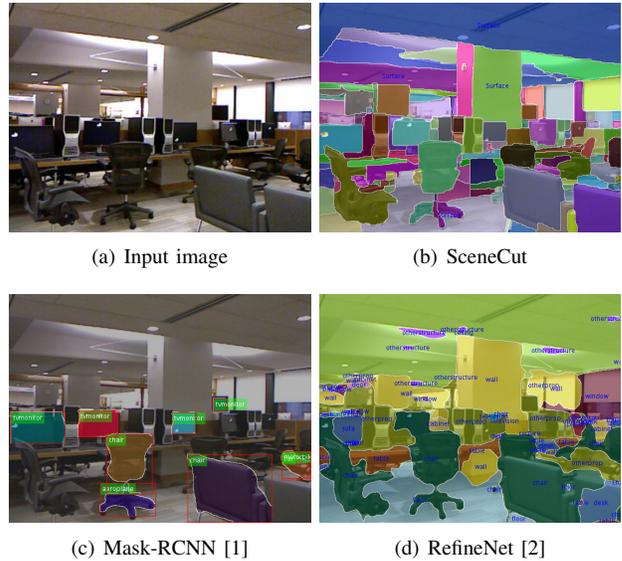

**Fig. 1:** SceneCut v.s. Mask-RCNN [1] v.s. RefineNet [2]. SceneCut segments the input image into individual class-agnostic objects and surfaces, encoded by different colors. RefineNet is unable to separate object instances while Mask-RCNN is unable to detect surfaces and unknown objects. Best viewed on-screen.

## I. INTRODUCTION

In recent years, scene understanding driven by multi-class semantic segmentation [2], or object detection [1], [3] has progressed significantly thanks to the power of deep learning. However, a major limitation of these deep learning based approaches is that they only work for a set of known object classes that are used during training. In contrast, autonomous robots often operate under *open-set* conditions [4] in many application domains, i.e. they will inevitably encounter objects that were not part of the training dataset. State-of-the-art methods such as Mask-RCNN [1], YOLO9000 [3] fail to detect such unknown objects. This behavior is detrimental to the goal of robotic scene understanding that would ideally result in a semantically meaningful map [5]–[8] comprising all objects, environmental structures, and their various complex relations. The ability to extract information about objects (e.g., semantic classes and affordances [9]) and the scene geometry in complex environments under realistic, open-set conditions is increasingly important for robotics.

We propose SceneCut, a novel approach to segment a scene into class-agnostic objects without requiring semantic class annotations for training. Using a single RGB-D image, SceneCut can discover unseen objects in highly cluttered indoor environments, thus allowing a higher degree of completeness for robotic scene understanding or object-based mapping or SLAM [7], [8]. In contrast to state-of-the-art pixel-wise semantic segmentation [2], our approach can differentiate between different object *instances*, e.g., chair A and chair B. In addition, SceneCut reasons jointly over objects and scene geometry and therefore produces a segmentation of non-object surfaces such as ceilings and walls, as well as supporting surfaces such as floor or tabletops. Unlike methods [5], [10]–[12] that deal with geometric segmentation and object segmentation in isolation, our method simultaneously segments objects and planes of a scene in an unified formulation. SceneCut therefore ensures the segmentation of geometric surfaces is consistent with the discovered objects. It avoids typical segmentation errors and inconsistencies that are often observed in other methods such as when pictures are assigned to a wall, or mouse pads are assigned to a desk or tabletop segment. Fig. 1 illustrates a typical cluttered indoor scene and compares the segmentation produced by SceneCut with RefineNet [2] and Mask-RCNN [1].

[1]The authors are with the Australian Centre for Robotic Vision.
[2]Trung T. Pham, Thanh-Toan Do, and Ian Reid are with the University of Adelaide, SA 5005, Australia.
[3]Niko Sünderhauf is with Queensland University of Technology (QUT), Brisbane, QLD 4001, Australia.
Contact: trung.pham@adelaide.edu.au
The authors acknowledge the support of the Australian Research Council through the Centre of Excellence for Robotic Vision (CE140100016) and Laureate Fellowship (FL130100102) to IDR.

A key aspect of our approach is a novel unified energy function that jointly quantifies geometric goodness-of-fit and objectness measure. The function, however, is difficult to optimize as its domain is over a continuous space of plane parameters and a discrete collection of image regions. For tractability, we utilize hierarchical segmentation trees

[13] to sample potential object and surface candidates. In particular, the input image is converted to a region hierarchy, where nodes of the tree represent meaningful regions — potentially objects and surfaces. One interesting property of the segmentation tree is that a cut through the tree will result in a *flat* segmentation (i.e., one pixel belongs to a unique region). The problem becomes finding a tree cut that optimizes the geometric and object fitting function. As the problem has a tree-structure, the solution can be found efficiently and exactly using dynamic programming.

A vital ingredient in building a high-quality segmentation tree is accurate boundaries. We make use of the boundary map predicted by Convolutional Oriented Boundary (COB) network of [14]. The COB network has shown its superior performance over traditional boundary detection methods (e.g., [15]) that use local features such as colors and depths. Importantly, training such a boundary detection networks does not rely on semantic class information, and thus the model generalizes to unknown environments with unknown objects, as we will show in Section IV.

We evaluate the effectiveness of SceneCut on the indoor scene segmentation task using NYU [12] and TUM [16] datasets. Experimental results on NYU dataset reveals that SceneCut greatly outperforms state-of-the-art image segmentation methods. In addition, using the TUM dataset we show that SceneCut works well on different environments without retraining or re-tuning hyper-parameters.

Our paper is structured as follows. We begin by reviewing related work in Section II. In Section III we describe the details of SceneCut. In Section IV we investigate the performance of the SceneCut method. Finally, we conclude and highlight some limitations in Section V.

## II. RELATED WORK

Simultaneous Localization and Mapping (SLAM) is a key component in many service robots that need to map new unstructured environments and navigate within them automatically. Advanced robots, however, require more sophisticated interactions with the environments, thus need to understand the world around them in terms of semantic entities. Noticeably, Salas-Moreno et. al. introduced SLAM++ [7] which simultaneously performs object recognition and camera tracking. Semantically-enriched maps with object entities are then constructed. In [17], McCormac et. al. proposed SemanticFusion which labels 3D reconstructed maps with semantic object classes. This method fuses semantic labels in real-time along with scene reconstruction. Recently, Sünderhauf et. al. [8] developed a method for object-oriented semantic mapping, where individual object instances are key entities in the maps. Nevertheless, these methods require either 3D object models available prior to execution (e.g., in [7]) or object detectors trained previously (e.g., in [8]). These systems face difficulties when deployed in unknown environments. One possible solution is to jointly perform object discovery, camera tracking and mapping, such as the method in [18], which relies on the unsupervised segmentation method [10] to segment planes and objects from incoming RGB-D frames.

However, the segmentation method [10] performs badly on general messy and cluttered indoor environments, as shown in Section IV. Inspired by this research trend, in this work, we aim at providing a robust joint geometric and object segmentation method, which will be useful for existing object SLAM or semantic mapping systems such as [8], [18].

Our work is closely related to unsupervised segmentation methods which are commonly used used for object discovery. Felzenszwalb and Huttenlocher [19] proposed a graph-based approach to color image segmentation. A graph structure is imposed on the image, and edge weights are computed based on color differences. A greedy merging algorithm is applied to the graph to obtain final segmentations. Karpathy et. al. [20] used this method to discover objects in 3D, where the graph is constructed over the point cloud, and edge weights are computed using colors and normals. Similarly, Trevor et. al. [10] defined binary edge weights for plane extraction using RGB-D images. To discover objects, the method in [10] first removes points belonging to detected planes, then segments remaining points into objects using Euclidean clustering. Pham et. al. [11] proposed a novel constrained plane segmentation method for 3D scenes. Unlike these methods which decouple geometric and semantic segmentation, SceneCut jointly segments surfaces and discovers object instances in an unified formulation.

One of the limitations of unsupervised segmentation methods is that these methods often require tuning a parameter(s) to control the number of regions in the segmentations. In practice, there is no single threshold or parameter that works well for all objects and different images. Hierarchical segmentation methods [13], [21] tackle this issue by output, instead, multiple segmentations nested in a tree so that hopefully objects are correctly segmented at some level of the tree. Finding correct levels for different objects is, however, non-trivial. The method in [22] attempted to align the scales so that all objects are at the same scale/level. SceneCut greatly makes use of hierarchical segmentation trees not only for finding correct object regions, but also for sampling the space of geometric plane models. Instead of realigning the object scales as in [22], we find the best object regions and surface models by searching for the optimal tree cut that maximizes a joint geometric and object fitting function.

Our work also bears a resemblance to the works of Gould et. at. [23] and Silberman et. al [12]. In [23], the authors proposed to a learning based model to decompose an outdoor scene into a ground plane and semantic-class regions. The method in [12] tried to parse an indoor scene into semantic and surface regions, as well as estimate support relations. Similar to SceneCut, the inference also makes use of hierarchical segmentation tree, but assumes that 3D scene structure has been accurately estimated using RANSAC. Furthermore, the inference is based on a linear programming formulation which is expensive and does not scale well with the nodes of the tree. In contrast, our geometric and object fitting function can be optimized efficiently using dynamic programing. Importantly, these two methods [12], [23] face difficulties when encountering objects that were not seen during training.

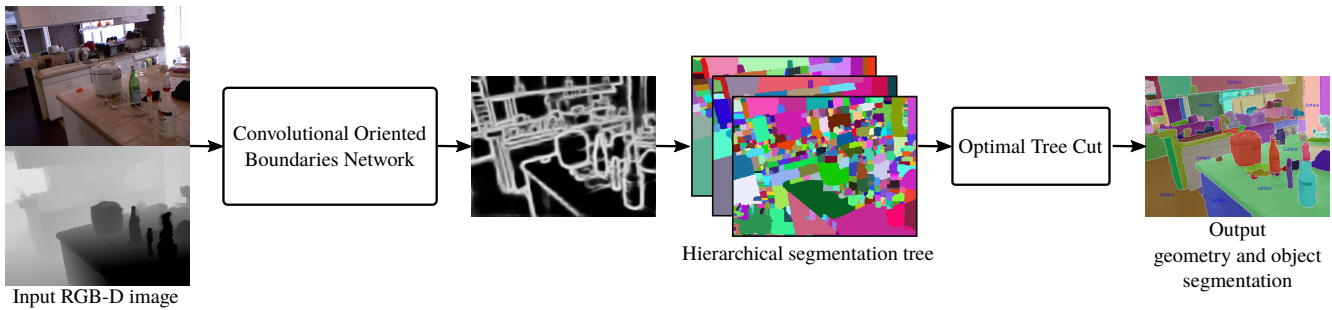

**Fig. 2:** An overview of SceneCut. The input image is first passed to a deep convolutional neural network [14] to predict a boundary map, which is then used to construct a hierarchical segmentation tree. Nodes of the tree serve as potential object and surface candidates. The optimal tree cut is optimized to output the final geometric and object segmentation.

## III. SCENECUT

### A. Overview

Given an input image, our goal is to decompose the image into meaningful non-overlapping regions, where each region represents either an object (e.g., chair, computer) or a planar surface (e.g., wall, floor, tabletop) instance. Figure 2 demonstrates the overview of our approach. The input image is first passed to a deep convolutional neural network to predict a boundary map, which is used to construct a hierarchical segmentation tree. We then search for an optimal tree cut that maximizes a geometric and object fitting function. The found optimal tree cut yields the final segmentation.

### B. Joint Geometry and Object Segmentation

We denote the input RGBD image as $\mathcal{I} = \{\mathbf{x}_i\}$. The task is to partition the image $\mathcal{I}$ into non-overlap regions $\mathcal{S} = \{S_1, S_2, \ldots, S_N\}$ such that each region represents an object or a planar surface instance. We also estimate the corresponding region parameters $\Theta = \{\theta_1, \theta_2, \ldots, \theta_N\}$, where $\theta$ can be either plane parameters of a surface or an integer indicating object identity (not object class). Effectively, $\theta_i$ can be seen as the "label" of the region $S_i$. Note that multiple different regions in $\mathcal{S}$ might have the same label. This property is useful for handling occlusion — for instance the ground floor is fragmented into several disconnected visible regions in the image due to objects such as chairs, tables resting on it.

To find the optimal segmentation $\mathcal{S}$ and its corresponding parameters $\Theta$, we need a way to assess its quality. Assume that a function $\psi(S_i, \theta_i) : [S_i, \theta_i] \to \mathbb{R}$ is provided to measure the score (energy) of the region $S_i$ with parameters $\theta_i$ being an object or a surface. The quality of $\{\mathcal{S}, \Theta\}$ can be computed as:

$$E(\mathcal{S}, \Theta) = \sum_{i=1}^{N} \psi(S_i, \theta_i). \quad (1)$$

The tricky part is to design the function $\psi(S_i, \theta_i)$ that properly calibrates geometric goodness-of-fit and objectness quantity. In this work, we use a sum of log likelihood function:

$$\psi(S_i, \theta_i) = \begin{cases} \sum_{\mathbf{x} \in S_i} \log g(\mathbf{x}, \theta_i) & \text{if } \theta_i \text{ represents a plane} \\ |S_i| \log f(S_i) & \text{otherwise,} \end{cases} \quad (2)$$

where $f(S_i) : S_i \to [0, 1]$ measures the likelihood of region $S_i$ being an object, $g(\mathbf{x}, \theta_i) : \mathbf{x}, \theta_i \to [0, 1]$ computes the likelihood of assigning pixel $\mathbf{x}$ to the plane model $\theta_i$, $|S_i|$ is the size (number of pixels) of the region $S_i$.

The problem of surface and object segmentation becomes finding the optimal segmentation $\mathcal{S}^*$ and parameters $\Theta^*$ such that

$$\{\mathcal{S}^*, \Theta^*\} = \underset{\mathcal{S}, \Theta}{\operatorname{argmax}} \, E(\mathcal{S}, \Theta). \quad (3)$$

Nevertheless, solving the maximization (3) is non-trivial due to the complex search spaces of $\mathcal{S}$ and $\Theta$. Next, we will show how to approximate the optimization (3) using hierarchical segmentation tree and tree cuts.

### C. Hierarchical Segmentation Tree

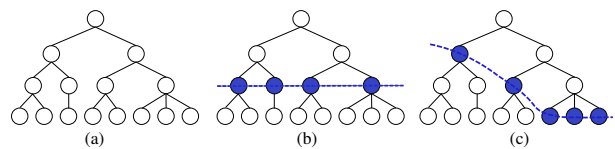

**Fig. 3:** An example of a segmentation tree (a) and different cuts. (b) a simple horizontal cut. (c) an optimal cut.

Hierarchical segmentation represents an image at multiple scales, where similar pixels are grouped into region hierarchies. By construction, nodes of the tree present *meaningful things* appearing in the image. In this work, we leverage region hierarchies to approximate the space of scene surfaces and object candidates.

A common way to construct a segmentation tree for an input image is based on a boundary map estimated from the image. The boundary map can be converted to a region hierarchy using the Ultrametric Contour Map (UCM) [21]. Traditionally, the boundaries is computed using local features such as colors and depths. Recent works (e.g., [14]) show that high-quality boundary maps can be predicted directly from images using deep convolutional networks. In this work, we resort to the COB network [14] for the boundary prediction. As we are interested in indoor scenes with RGBD data, we select the COB model previously trained on NYU dataset [12]. We will show that this model still works well on many other datasets different from NYU, such as TUM [16].

Let $\mathcal{T}$ be a hierarchical segmentation tree obtained from the image $\mathcal{I}$, comprising $K$ nodes, i.e., $\mathcal{T} = \{R_1, \ldots, R_K\}$. Each node represents a single image region, and the root node is the whole image. Moreover, each node has one and only one parent (other than the root) and has at least one child (other than the leaves). We consider each image region (node of the tree) as an object candidate. We also fit a plane model to 3D points of each region, resulting in a set the potential plane models $\mathcal{P} = \{\mathbf{p}_1, \mathbf{p}_2, \ldots, \mathbf{p}_K\}$.[1]

One interesting property of the segmentation tree is that a "tree cut" will result in a flat image segmentation $\mathcal{S}$. Figure 3 illustrates a segmentation tree and different tree cuts. Let $\mathcal{C}$ be the space of all possible tree cuts, it is clear that $\mathcal{S} \in \mathcal{C}$. In the next section, we will show to maximize the segmentation function (1) over the spaces of $\mathcal{C}$ and $\mathcal{P}$ exactly using dynamic programming.

### D. Optimal Tree Cut using Dynamic Programming

We denote $\mathcal{Y} = \{0, 1, 2, \ldots, K, \mathbf{p}_1, \mathbf{p}_2, \ldots, \mathbf{p}_K\}$ as a joint set of object and planar surface candidates. Finding the best tree cut can be reformulated as optimizing label variables $\mathcal{L} = [l_1, l_2, \ldots, l_K]$, where $l_i$ takes value in $\mathcal{Y}$. More specifically, for each node $R_i$ of the tree $\mathcal{T}$, we introduce a label $l_i$ — if $l_i$ is zero, node $R_i$ is not selected, otherwise node $R_i$ is included in the solution $\mathcal{S}$. Furthermore, $l_i$ will take value $\mathbf{p}_i$ if region $R_i$ is a planar surface. Now the objective function (1) becomes:

$$E(\mathcal{T}, \mathcal{L}) = \sum_{R_i \in \mathcal{T}} \psi(R_i, l_i), \qquad (4)$$

where

$$\psi(R_i, l_i) = \begin{cases} \sum_{\mathbf{x} \in R_i} \log g(\mathbf{x}, l_i) & \text{if } l_i \in \mathcal{P} \\ |R_i| \log f(R_i) & \text{if } l_i = i \\ -\infty & \text{otherwise.} \end{cases} \qquad (5)$$

Notice that multiple surface regions can take the same label (plane parameters) in $\mathcal{P}$, therefore disconnected surface regions can be merged. However, the current formulation is unable to merge isolated regions belonging to the same object (which can arise due to occlusion).

Given $\mathcal{L}$, we can obtain $\mathcal{S} = \{R_i \mid l_i \neq 0\}$. However, not any labelling $\mathcal{L}$ results in a valid tree cut. In order for $\mathcal{S}$ (obtained from $\mathcal{L}$) to be a valid tree cut, if region $R_i$ has label $l_i \neq 0$, its children and parent nodes will be assigned to label 0. Assume the root of the tree is $R_1$, the maximum of the function (4) with the "tree cut" constraint can be obtained recursively as follow:

$$\max E(\mathcal{T}, \mathcal{L}) = \max(\psi(R_1, l_1), E(\mathcal{T}_{R_1}, \mathcal{L}_{\mathcal{T}_{R_1}})), \qquad (6)$$

where $\mathcal{T}_{R_1}$ is the subtree rooted at node $R_1$, and $\mathcal{L}_{\mathcal{T}_{R_1}}$ is the corresponding labels for the subtree nodes.

Thanks to the tree-structure property, we can find the optimal $\mathcal{L}^*$ using dynamic programming as follow. The optimization includes one forward and one backward pass.

[1] In practice, planes with less than 5000 inliers or smaller than $0.5m^2$ are discarded.

The forward pass, proceeding from the bottom to the top of the tree, computes energies at each node and its subtree. The backward pass will compare, from the top to bottom, the energies of the current node and its subtree to obtain the optimal labels. This dynamic programming method is highly efficient with complexity $O(K)$, where $K$ is the total number of nodes.

### E. Objectness measure

For each region of the segmentation tree, we compute its *objectness* $f(.)$ — the likelihood of being an object. A common way to measure objectness is to analyze object size, shape and appearance, e.g., [20]. However, indoor objects' sizes, shapes and appearances vary significantly, making it difficult to define a reliable objectness measure. Instead, we compute the objectness measure using boundary strengths. Intuitively, a region with strong external boundary and weak internal boundary is more likely an object. Furthermore, regions at higher levels are more likely under-segmented, and vice versa regions at lower levels are more likely over-segmented. Given a region $R$ at level $l_R \in [0, 1]$, its external boundary score $b_R^{ex}$ is the highest level where it starts merging with other regions, its internal boundary score $b_R^{in}$ is the lowest level of its children. The objectness of $R$ is computed as:

$$f(R) = |b_R^{ex} - b_R^{in}| \exp(-\frac{(l_R - l_{mid})^2}{\sigma_l^2}), \qquad (7)$$

where $l_{mid}$ represents a "mean" level, and the degree of deviation from $l_{mid}$ is controlled by parameter $\sigma_l$. We found that regions at levels higher than 0.6 are mostly under-segmented, thus we set $l_{mid} = 0.3$ and $\sigma_l = 0.2$ in all our experiments.

### F. Goodness-of-fit

The likelihood of a pixel belonging to a planar surface is computed using several criteria. First, its 3d location must lie close to the surface. Second, its normal should agree with the normal of the surface. Lastly, the pixel color should not be much different from the mean color of the surface. The last criterion is very useful to deal with flat objects, e.g., pictures attached to the wall. Given a pixel $\mathbf{x}$ with color $\mathbf{x}_c$, 3d location $\mathbf{x}_l$ (in homogeneous coordinate) and normal $\mathbf{x}_n$, a plane model with parameters $\mathbf{p}$, normal $\mathbf{p}_n$ and mean color $\mathbf{p}_c$, the likelihood of assigning $\mathbf{x}$ to plane $\mathbf{p}$ is:

$$g(\mathbf{x}, \mathbf{p}) = \exp(\frac{-(\mathbf{x}_l \cdot \mathbf{p})^2}{\sigma_d^2}) \times \exp(\frac{-(1 - \mathbf{x}_n \cdot \mathbf{p}_n)^2}{\sigma_n^2}) \quad (8)$$
$$\times \exp(\frac{-\|\mathbf{x}_c - \mathbf{p}_c\|^2}{\sigma_c^2}),$$

where $(\mathbf{x}_l \cdot \mathbf{p})^2$ is the squared distance from 3D point $\mathbf{x}_l$ to plane $\mathbf{p}$, $(\mathbf{x}_n \cdot \mathbf{p}_n)$ is the dot product between normal vectors, $\sigma_d, \sigma_c$ and $\sigma_n$ are variance parameters. In Eq.(8) we assume $\mathbf{x}_c$ and $\mathbf{p}_c$ are scalars, though vectors can also be applied. We use $\sigma_d = 0.02$ (in meters), $\sigma_n = 0.3$ and $\sigma_c = 0.1$ (for HSV colors) in all the experiments.

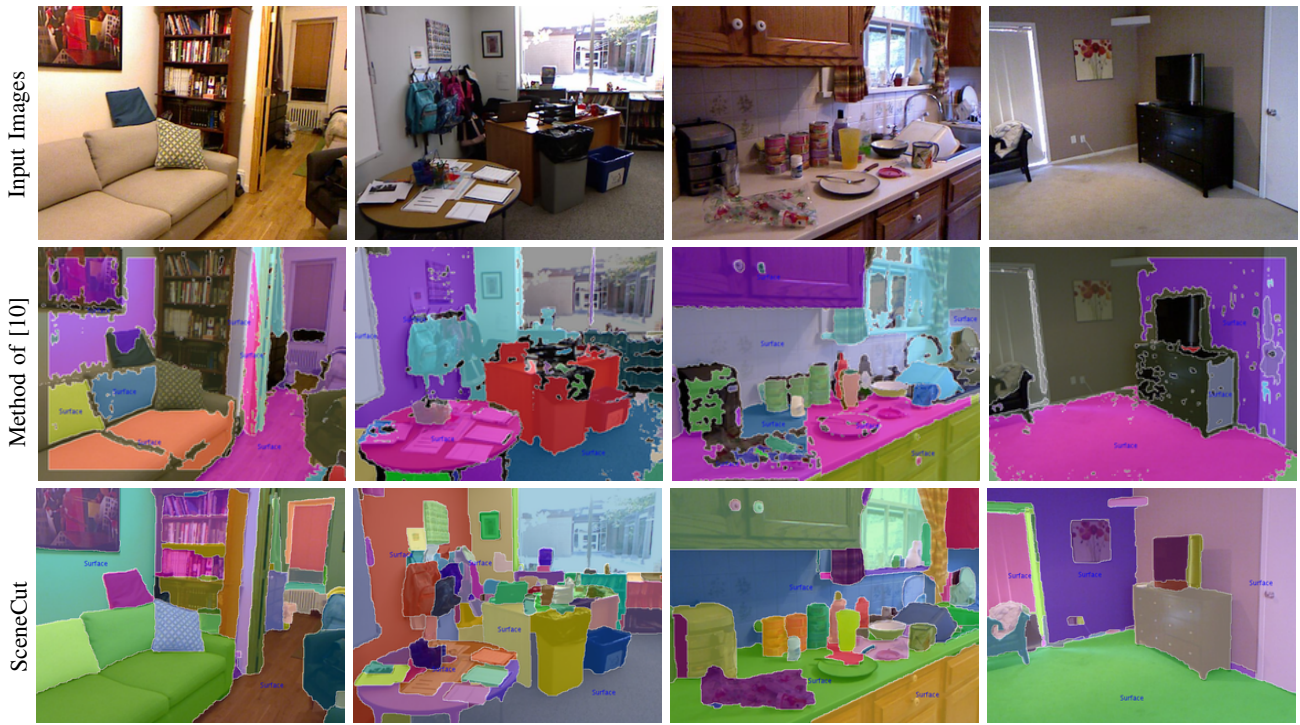

**Fig. 4:** Sampled segmentation results of SceneCut and [10] using images from NYU dataset. Best viewed on-screen.

## IV. EXPERIMENTS

### A. Datasets and Baselines

We use indoor RGB-D NYU [12] and TUM [16] datasets to evaluate the effectiveness of SceneCut. The NYU dataset contains 654 RGB-D testing images with ground-truth semantic-object annotations, which is widely used for evaluating indoor semantic instance segmentation. However, as our goal is class-independent object segmentation, we do not use semantic classes for evaluation. The TUM dataset, on the other hand, has no ground-truth object annotations as it was created for evaluating camera tracking and scene reconstruction. Nevertheless, this dataset can be viewed as unknown environments with unknown objects, which is useful for testing the object discovery capability. We select three sequences from the TUM repository, namely freiburg1 desk2, freiburg2 desk, and fr3 household, and randomly sample 600 frames for testing. Since there is no ground-truth annotations, we only provide qualitative evaluations on this dataset.

Since our method SceneCut aims at reasoning about surfaces and discovering object instances in indoor scenes, our closest competitors include: 1) the method of [10] which first extracts planes using connected components, and then discovers objects using Euclidean clustering, 2) the method of Silberman et al. [12] which first infers the overall 3D scene structure (in terms of planes) using RANSAC, then separates objects and reasons support relations via an integer programming formulation. We also compare SceneCut against popular unsupervised image segmentation approaches including LCCP [24] and graph-based segmentation [19]. Unlike SceneCut, these unsupervised segmentation methods only decompose images into meaningful regions without knowing their object or non-object labels.

The NYU dataset does not provide accurate ground-truth information for geometric surfaces although the dataset has annotated ground-floors, walls and ceilings, but other objects such as table tops or cabinet tops can be considered as planar surfaces too. Therefore, we evaluate the performance using class-independent segmentation accuracy without considering geometric or object labels. We use two popular measures: (symmetric) segmentation covering (SSC) and F measure for regions [25].

### B. Results

Table I reports the segmentation results of all the considered methods using NYU dataset. It is clear that SceneCut significantly outperforms its competitors that solve geometric and object segmentation independently. SceneCut is also much more accurate than the competing unsupervised segmentation methods [19], [24]. The superiority of SceneCut over these methods is due to two reasons: 1) joint reasoning about objects and surfaces using optimal tree cut and 2) the use of deep learning to predict high-quality boundaries. Figure 4 visually demonstrates segmenatation results. In comparison with other methods, SceneCut shows clear improvements. For example, the method [10] merges wall and picture into a single surface. Touching objects (e.g., table and trash bins) are also not segmented correctly.

To demonstrate that the improvement is not from the use of deep boundary prediction only, we evaluate the performance of the greedy horizontal cut (see Figure 3) with the best dataset scale (BDS) that is computed using ground truth information. Note that the best estimated scale of the NYU dataset might not translate well to different unknown environments. As reported in Table I, our optimal tree cut is still better than the best horizontal cut. We further inspect the cases where SceneCut is quantitatively less accurate than the best horizontal cut. Surprisingly, we find that our segmentation results are qualitatively better. As shown in Figure 5, it can be observed that SceneCut correctly segments multiple planar surfaces in the scenes while the ground-truth annotations show a single wall and a single cabinet region. This further demonstrates the strong capability of SceneCut for joint geometric and object segmentation.

| Method | Use depth | Learning | Surface | Object | SSC | F region |
|---|---|---|---|---|---|---|
| Silberman et al. [12] | ✓ | ✓ | ✓ | ✓ | 61.10 | - |
| LCCP [24] | ✓ | ✗ | ✗ | ✓ | 57.60 | - |
| Graph-based Segmentation [19] | ✗ | ✗ | ✗ | ✓ | 56.56 | 62.42 |
| Connected Components [10] | ✓ | ✗ | ✓ | ✓ | 51.34 | 55.93 |
| **SceneCut** (proposed approach) | ✓ | ✓ | ✓ | ✓ | **75.08** | **80.28** |
| BDS horizontal cut (0.3) | ✓ | ✓ | ✗ | ✓ | 74.72 | 80.19 |

TABLE I: Quantitative comparison results on 654 NYU RGB-D testing images. The results of LCCP and Silberman et al. [12] are taken from [24] and [12] respectively.

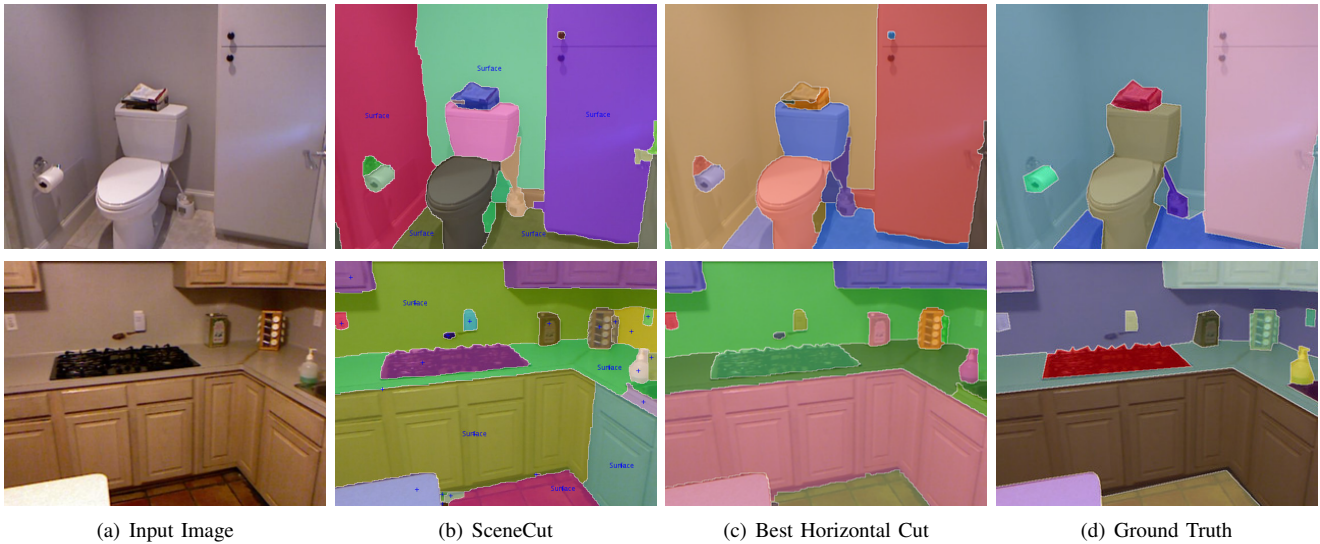

(a) Input Image    (b) SceneCut    (c) Best Horizontal Cut    (d) Ground Truth

**Fig. 5:** Typical examples where SceneCut is quantitatively less accurate than horizontal cut (compared to the ground truth). However, SceneCut correctly segments the surfaces. Best viewed on-screen.

Figure 6 visualizes the segmentation results of SceneCut and its competitor [10] using images from the TUM dataset, where none of the methods was trained on. On these unknown environments, SceneCut still achieves good results and is qualitatively better than [10] . Notice that the TUM dataset contains objects such as teddy bear, cubic, sealing tape, etc., that never appear in the NYU dataset which was used for training the boundary network. Nevertheless, these objects are still segmented correctly by SceneCut. On the other hand, SceneCut fails to segment some objects precisely. For instance, the chair is over-segmented into two components: a chair back and a chair seat. Such a segmentation, however, is also useful as these object parts themselves can be considered as individual objects to some extent.

### C. Runtime Analysis

Except for the boundary prediction and segmentation tree construction making use of C++ and GPU implementation[2], other steps of SceneCut including computing objectness, plane parameters and tree cut are run using unoptimized Matlab implementation. Table II presents average runtimes for each step. There are several ways to reduce the computation cost. For instance, in the current implementation, features of image regions (e.g., region areas) are computed independently. This computation will be much cheaper if properly exploiting the segmentation tree, i.e., propagating the features from leaf nodes to parent nodes. Similar strategy can be applied for fitting planes to depth regions. Moreover, a proper tree data structure in C++ can significantly reduce the computation cost of finding optimal tree cuts.

| Boundary | Seg. Tree | Objectness | Plane Estimation | Tree Cut |
|---|---|---|---|---|
| 0.28 | 0.51 | 1.75 | 1.24 | 1.26 |

**TABLE II:** Average running times for each step of SceneCut (in second).

---

[2]https://github.com/kmaninis/COB

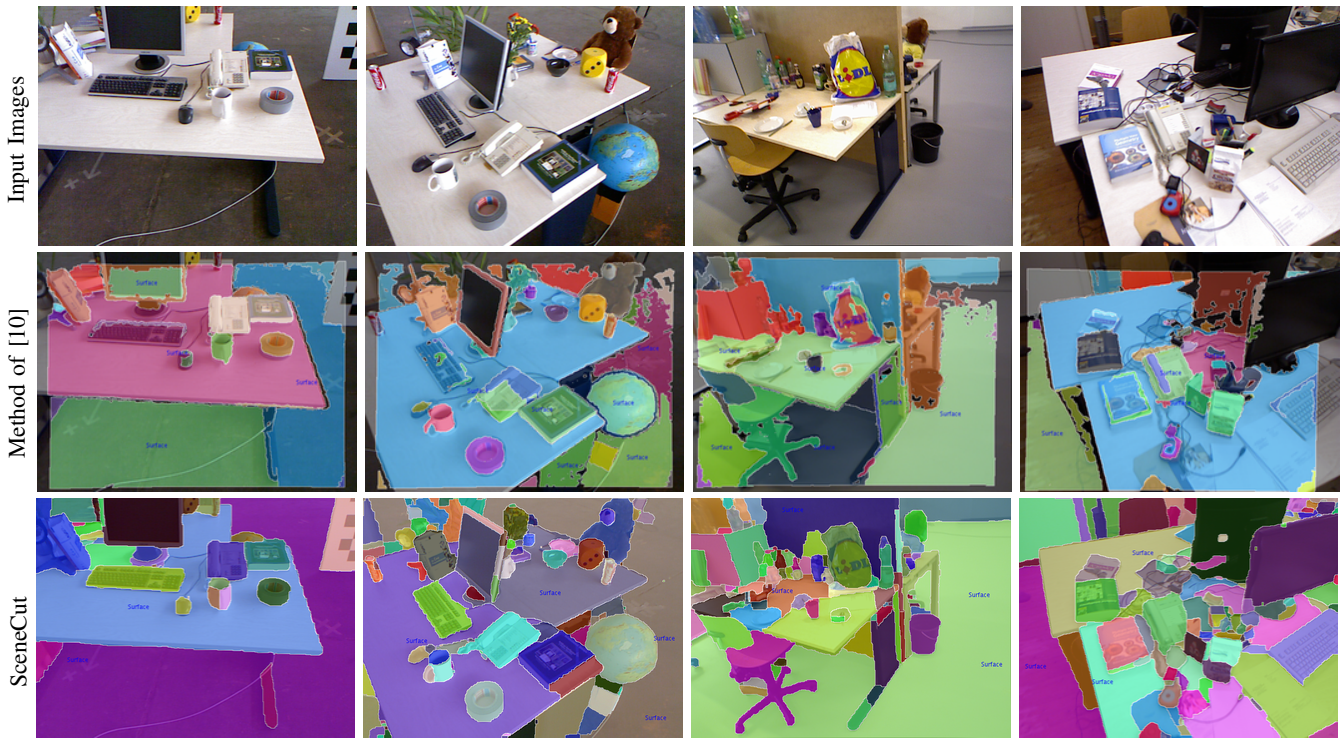

**Fig. 6:** Sampled segmentation results of SceneCut and [10] using images from TUM dataset. Best viewed on-screen.

## V. CONCLUSIONS

We have presented SceneCut that has a capability to simultaneously infer scene geometry and discover objects of indoor scenes. Such a capability is useful for autonomous robots working in *open-set* conditions [4], where the robots will unavoidably encounter novel objects that were not part of the training dataset. SceneCut automatically estimates supporting planar surfaces and discovering objects using an unified formulation. In particular, we proposed an unified energy function that jointly encodes geometric fitting and objectness measures. We show how to efficiently maximize the energy by utilizing hierarchical segmentation trees and tree cuts. Moreover, the method enjoys the power of deep learning applied for predicting object boundaries directly from input images. High-quality boundary prediction has led to highly accurate geometric and object segmentation. The experimental results demonstrate that our proposed approach greatly outperforms many existing RGB-D scene segmentation methods.

A limitation of the current approach is its inability to correctly segment occluded object instances that are fragmented into disconnected regions in the image. Also the hierarchical segmentation trees sometimes fail to merge regions belonging to the same object instance due to badly estimated boundaries. One way to work around is to use higher-order parametric models such as ellipsoid or NURB to model objects rather simple image regions considered in the current approach.

## REFERENCES

[1] K. He, G. Gkioxari, P. Dollár, and R. Girshick, "Mask R-CNN," in *Proceedings of the International Conference on Computer Vision (ICCV)*, 2017.

[2] G. Lin, A. Milan, C. Shen, and I. Reid, "RefineNet: Multi-path refinement networks for high-resolution semantic segmentation," in *CVPR*, Jul. 2017.

[3] J. Redmon and A. Farhadi, "YOLO9000: better, faster, stronger," 2017.

[4] W. J. Scheirer, A. de Rezende Rocha, A. Sapkota, and T. E. Boult, "Toward open set recognition," *IEEE Transactions on Pattern Analysis and Machine Intelligence*, vol. 35, no. 7, pp. 1757–1772, 2013.

[5] T. T. Pham, I. Reid, Y. Latif, and S. Gould, "Hierarchical higher-order regression forest fields: An application to 3d indoor scene labelling," in *The IEEE International Conference on Computer Vision (ICCV)*, December 2015.

[6] C. Cadena, L. Carlone, H. Carrillo, Y. Latif, D. Scaramuzza, J. Neira, I. Reid, and J. J. Leonard, "Past, present, and future of simultaneous localization and mapping: Toward the robust-perception age," *IEEE Transactions on Robotics*, vol. 32, no. 6, pp. 1309–1332, 2016.

[7] R. F. Salas-Moreno, R. A. Newcombe, H. Strasdat, P. H. J. Kelly, and A. J. Davison, "Slam++: Simultaneous localisation and mapping at the level of objects," in *2013 IEEE Conference on Computer Vision and Pattern Recognition*, June 2013, pp. 1352–1359.

[8] N. Sünderhauf, T.-T. Pham, Y. Latif, M. Milford, and I. D. Reid, "Meaningful maps with object-oriented semantic mapping," in *IEEE/RSJ International Conference on Intelligent Robots and Systems*, 2017.

[9] T.-T. Do, A. Nguyen, I. D. Reid, D. G. Caldwell, and N. G. Tsagarakis, "Affordancenet: An end-to-end deep learning approach for object affordance detection," in *IEEE International Conference on Robotics and Automation (ICRA)*, 2018.

[10] A. J. B. Trevor, S. Gedikli, R. B. Rusu, and H. I. Christensen, "Efficient organized point cloud segmentation with connected components," 2013.

[11] T. T. Pham, M. Eich, I. D. Reid, and G. Wyeth, "Geometrically consistent plane extraction for dense indoor 3d maps segmentation," in *IEEE/RSJ International Conference on Intelligent Robots and Systems*, 2016, pp. 4199–4204.

[12] N. Silberman, D. Hoiem, P. Kohli, and R. Fergus, "Indoor segmentation and support inference from rgbd images," in *Proceedings of the 12th European Conference on Computer Vision - Volume Part V*, ser. ECCV'12. Berlin, Heidelberg: Springer-Verlag, 2012, pp. 746–760.

[13] P. Arbelaez, M. Maire, C. Fowlkes, and J. Malik, "Contour detection and hierarchical image segmentation," *IEEE Transactions on Pattern*

*Analysis and Machine Intelligence*, vol. 33, no. 5, pp. 898–916, May 2011.
[14] K. Maninis, J. Pont-Tuset, P. Arbeláez, and L. V. Gool, "Convolutional oriented boundaries: From image segmentation to high-level tasks," *IEEE Transactions on Pattern Analysis and Machine Intelligence (TPAMI)*, 2017.
[15] S. Gupta, P. Arbelez, and J. Malik, "Perceptual organization and recognition of indoor scenes from rgb-d images," in *2013 IEEE Conference on Computer Vision and Pattern Recognition*, June 2013, pp. 564–571.
[16] J. Sturm, N. Engelhard, F. Endres, W. Burgard, and D. Cremers, "A benchmark for the evaluation of rgb-d slam systems," in *Proc. of the International Conference on Intelligent Robot Systems (IROS)*, Oct. 2012.
[17] J. McCormac, A. Handa, A. Davison, and S. Leutenegger, "Semanticfusion: Dense 3d semantic mapping with convolutional neural networks," in *2017 IEEE International Conference on Robotics and Automation (ICRA)*, May 2017, pp. 4628–4635.
[18] S. Choudhary, A. J. B. Trevor, H. I. Christensen, and F. Dellaert, "Slam with object discovery, modeling and mapping," *2014 IEEE/RSJ International Conference on Intelligent Robots and Systems*, pp. 1018–1025, 2014.
[19] P. F. Felzenszwalb and D. P. Huttenlocher, "Efficient graph-based image segmentation," *International Journal of Computer Vision*, vol. 59, no. 2, pp. 167–181, Sep 2004.
[20] A. Karpathy, S. Miller, and L. Fei-Fei, "Object discovery in 3d scenes via shape analysis," in *2013 IEEE International Conference on Robotics and Automation*, May 2013, pp. 2088–2095.
[21] P. Arbelaez, "Boundary extraction in natural images using ultrametric contour maps," in *2006 Conference on Computer Vision and Pattern Recognition Workshop (CVPRW'06)*, June 2006, pp. 182–182.
[22] Y. Chen, D. Dai, J. Pont-Tuset, and L. Van Gool, "Scale-aware alignment of hierarchical image segmentation," in *Computer Vision and Pattern Recognition (CVPR)*, 2016.
[23] S. Gould, R. Fulton, and D. Koller, "Decomposing a scene into geometric and semantically consistent regions." in *ICCV*. IEEE Computer Society, 2009, pp. 1–8.
[24] S. Stein, M. Schoeler, J. Papon, and F. Worgotter, "Object partitioning using local convexity," in *Computer Vision and Pattern Recognition (CVPR), 2014 IEEE Conference on*, June 2014, pp. 304–311.
[25] J. Pont-Tuset and F. Marques, "Supervised evaluation of image segmentation and object proposal techniques," *IEEE Transactions on Pattern Analysis and Machine Intelligence (TPAMI)*, vol. 38, no. 7, pp. 1465–1478, 2016.